\documentclass{bmvc2k}
\usepackage{multirow}
\usepackage{bm}
\usepackage{amsmath}
\usepackage{mathtools}
\usepackage{amssymb}
\usepackage{booktabs}
\usepackage{hyperref}
\usepackage[ruled,vlined,linesnumbered]{algorithm2e}

\title{Every Step of the Way: Video-based Parkinsonian Turning Step Counting}

\addauthor{Qiushuo Cheng}{qiushuo.cheng@bristol.ac.uk}{1}
\addauthor{Jingjing Liu}{jingjing.liu@bristol.ac.uk}{1}
\addauthor{Catherine Morgan}{catherine.morgan@bristol.ac.uk}{2}
\addauthor{Alan Whone}{alan.whone@bristol.ac.uk}{2}
\addauthor{Majid Mirmehdi}{majid@cs.bris.ac.uk}{1}
\addinstitution{
School of Computer Science\\
Faculty of Engineering\\
 University of Bristol\\
 Bristol, UK
}
\addinstitution{
North Bristol NHS Trust\\
Bristol, UK
}

\runninghead{Cheng et al}{Video-based Parkinsonian Turning Step Counting}

\begin{document}

\maketitle
\begin{abstract}
As a prominent symptom of Parkinson’s disease (PD), turning impairment is evaluated through parameters such as turning angle, duration, and particularly, the number of steps required to complete a turn, which directly reflects motor dysfunction.
Accurate step counting is challenging due to variability in real-world turning movements and atypical shuffling patterns in parkinsonian gait. Existing methods are predominantly wearable-based, requiring users to wear and manage dedicated devices, which can be inconvenient for continuous daily use.
To address this, we propose a passive, video-based framework that estimates step count in a coarse-to-fine manner using diverse motion representations. 
Specifically, an initial step count is estimated from foot movement signals derived from 3D human mesh recovery, providing high-level motion structures.
To incorporate fine-grained motion details, a motion encoder learns complementary gait dynamics from mesh and optical flow to refine the initial estimate. In this process, coarse foot movement signals query the pixel-level motion cues via cross attention to capture subtle parkinsonian gait dynamics.
To handle varying video lengths, we partition each video into clips and integrate clip-wise motion embeddings via multiple instance learning (MIL) for step count residual prediction.
Extensive experiments show our method consistently outperforms existing step counting methods on real-world PD turning datasets.
\end{abstract}

\section{Introduction} \label{sec: intro}

\begin{figure}[t]
    \centering
    \includegraphics[width=0.92\linewidth]{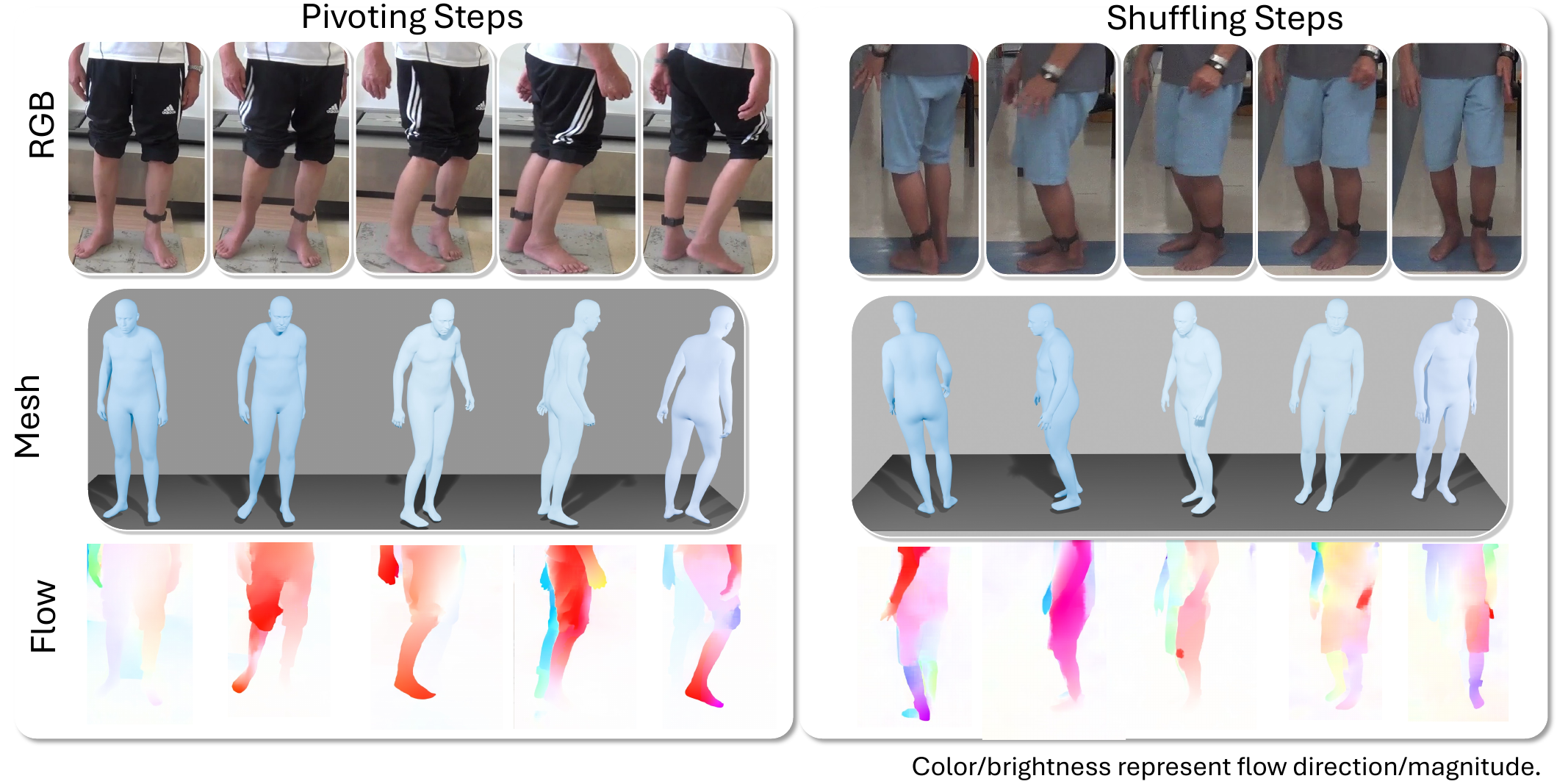}
    \caption{{Turning steps in Parkinson’s Disease (PD).} 3D mesh captures coarse-grained motion structure while optical flow preserves pixel-level motion details. (Left) One foot rotates around the other. (Right) Small, incremental steps.}
    \label{fig:concept}
\end{figure}

As one of the fastest-growing neurological disorders~\cite{21PDlancet}, Parkinson’s disease (PD) is characterised by motor symptoms including gait impairment, which is more pronounced~\cite{zampieri2010instrumented} during \textit{turning} compared to straight walking 
in early stages of the disease. Capturing and quantifying symptoms at this point is potentially vitally important for PD diagnosis, symptom evaluation and for clinical trials of novel therapeutics which are often targeted at prodromal and/or early stage disease. Furthermore, turning abnormalities can predispose to falls \cite{bloem2001prospective} and falls while turning increase the risk of hip fractures being sustained \cite{cummings1994falls}.
{Thus, quantifying {fundamental} turning parameters, such as turning duration~\cite{sensorlocation,lisa_turn_detection} or angle~\cite{turningangle,pham2017algorithm,cheng2025your}, 
plausibly represents an opportunity to predict falls and fractures in a person with PD.}
In particular, the {number of} steps taken to turn~\cite{morgan2022understanding} 
{can help distinguish between whether {a PD patient has taken} their medications or not, and correlates  strongly with the scores from the gold-standard clinical rating scale, the Movement Disorders Society-sponsored revision of the Unified Parkinson's Disease rating scale score (MDS-UPDRS)~\cite{updrs}. Furthermore, the number of turning steps can directly reflect core clinical manifestations of PD,} i.e. bradykinesia~\cite{14PDsubtype} and rigidity~\cite{2013rigid}, which can lead to {people with PD taking} an increased number of slow, short and shuffling steps~\cite{2015turndeficit, 19lancetPDgait} to complete turning {compared to people without PD} (see Figure \ref{fig:concept} (top row)).

Turning and steps are typically assessed within general clinical rating scales~\cite{tug, updrs} or measured using inertial measurement units (IMU)~\cite{turningangle,pham_imu_lowback}. However, manual ratings are labor-intensive and influenced by instructions and tester bias~\cite{19lancetPDgait}, while IMU-based approaches require managing wearable devices, which could be problematic for this population~\cite{2017wearableissues}. Moreover, their accuracy and reliability are highly sensitive to sensor placement~\cite{2017stepreview, sensorlocation2,sensorlocation}. While effective for general activity tracking~\cite{2025steplancet}, wearables may struggle to precisely measure the number of steps during specific activities. 
{Alternatively, pressure sensors can provide accurate step-level pressure changes but require {specialised} 
footwear~\cite{19reviewshoeinsole} or instrumented floors~\cite{17walkwaystepwearable}, which limits their applicability. }

In contrast, using video-based sensing enables passive monitoring and benefits from the widespread deployment of cameras in everyday environments, and given the widespread use of smartphones, could be used for clinical purposes in low-, middle- and high-income countries across the world. 
This work focuses on counting the number of steps from videos to help characterise impaired turning {in PD patients}.  
{The key challenge is that steps during turning 
in naturalistic settings especially for PD, are considerably more heterogeneous than those during straight walking, with various patterns~\cite{2017stepreview} (see Figure~\ref{fig:concept}) such as side-to-side, backward, pivoting, and small shuffling adjustments.} Steps also vary substantially both within and between individuals~\cite{morgan2022understanding} and across levels of disease severity~\cite{19lancetPDgait}. 
{However, existing video-based gait analysis methods~\cite{21gaitanalysis,24gaitanalysis,2024visiongaitanalysis} primarily focus on steps during general walking on treadmills or scripted walkways~\cite{gaitanalysis_Review}, with only few studies~\cite{21gaitanalysis,vision_step_during_tug} investigating turning steps of the scripted 180$^\circ$ turn during the Timed Up and Go (TUG) test.}
{These methods typically use human pose estimation to approximate 3D motion capture for predefined gait assessments,} which primarily aim to obtain precise biomechanical measurements such as joint angles or step length. Although such measurements can be used to infer step counts, they may not extend well to unconstrained free-living turning actions beyond laboratory settings.

In this paper, we introduce a coarse-to-fine framework for counting parkinsonian turning steps from videos. 
First, we utilise state-of-the-art human pose and shape models~\cite{gvhmr,prompthmr} to recover a 3D human mesh and construct a foot movement signal, from which an initial step count is obtained via peak detection~\cite{05peak}. While effective at capturing coarse motion structure {(Figure \ref{fig:concept} (middle row))}, the mesh alone is insufficient for the subtle and abnormal foot movements of PD, leading to artifacts such as footskate~\cite{footskate, HUMOR}. We therefore further refine the initial estimate via a motion encoder that fuses mesh-based signals with dense optical flow~\cite{poggi2025flowseek} through cross attention. Subtle foot movements are explicitly preserved in the spatially detailed and temporally sensitive flow dynamics {(Figure \ref{fig:concept} (bottom row))}, complementing the coarse mesh representation.
The refinement process is supervised on real-world PD turning videos of varying lengths using a multiple instance learning (MIL)~\cite{ilse2018mil} scheme. Each video is partitioned into clips, which are encoded as fused motion embeddings and aggregated via attention-based saliency scores to predict a video-level residual to compensate missing or extra steps.

{To demonstrate real-world applicability, we evaluate our approach on two clinically annotated turning datasets: PD-FOG~\cite{PDFOG} for in-clinic in-place turning assessment and Turn-REMAP~\cite{remap} for free-living turning at home.}  We conduct extensive experiments comparing against IMU-based algorithms~\cite{imu1,imu2,imu3}, gait analysis methods~\cite{gaitanalysis1,gaitanalysis2}, and end-to-end {repetitive} activity counting methods~\cite{repnet,ESCount}. Since the feature fusion process in our motion encoder is modality-agnostic, we further conduct multimodal ablations with additional inputs, including RGB and IMU signals.
We also validate the clinical relevance of our metrics by correlating with the gold standard MDS-UPDRS score~\cite{updrs}.

In summary, the contributions of this work are: (i) we present the first study on automatic step counting for parkinsonian turning action from videos, (ii) our proposed motion encoder fuses coarse mesh motion with fine-grained flow dynamics via cross attention to capture subtle parkinsonian foot movements, and 
(iii) extensive experiments on real-world PD datasets demonstrate that our approach outperforms existing vision-based methods {adapted for step count estimation}, such as ESCount~\cite{ESCount}, and notably, surpasses ankle-mounted IMU-based approaches such as Lucot et al.~\cite{imu2}, reducing the absolute step-count error from 37.7 to 15.9. 


\section{Related Works} \label{sec: related works}
{Conventional step definitions~\cite{gaitanalysis1, gaitanalysis2}, which assume a clearly identifiable swing phase with toe-off and heel strike, are not well suited to parkinsonian gait evaluation. Many irregularities of gait may be excluded, including shuffling gait with short step length~\cite{19lancetPDgait} and reduced foot clearance~\cite{2015turndeficit}, despite the fact that they are clinically informative~\cite{morgan2022understanding}. In our work, we consider any displacement of the foot during gait as important under the parkinsonian remit. Next, we consider works related to turning analysis in general, and those related to step counting with IMUs and/or vision sensors.}


\vspace*{2mm}

\noindent \textbf{{Quantifying Turning Impairment in PD --}}
{To assess turning abnormalities in PD, various technologies have been developed to quantify turning-related parameters. For example, clinical gait tests can be instrumented with IMUs~\cite{tug,turningangle,onshoe_stepturning,PDFOG} or analysed from video using human pose estimation~\cite{videoPDturninganalysis, tug_video} to measure turning duration, angle, and angular velocity. Moreover, pressure-sensitive insoles have been used to estimate the centre of pressure or centre of mass for assessing turning stability~\cite{insoles2} and balance~\cite{insoles}.
Some studies~\cite{turningangle,pham_imu_lowback, cheng2025your} further extend their method to turning actions during daily activities in more naturalistic settings.  In contrast to these approaches, our method focuses specifically on estimating step count from parkinsonian turning videos and is applicable in both clinical and free-living settings.}

\vspace*{2mm}

\noindent \textbf{Step Counting with Non-visual Sensors --}
Ever since the 15th century~\cite{2017stepreview}, human steps have been counted and measured by sensors for position tracking~\cite{20deadreckoning} or healthcare monitoring~\cite{2025steplancet}. Modern approaches~\cite{imu1,imu2,pham_imu_lowback,onshoe_stepturning, imu3} commonly rely on motion sensors such as inertial measurement units (IMU) embedded in wearable~\cite{2022step_variation} or mobile devices~\cite{2017mobile_tread}, which detect periodic step events by identifying characteristic patterns in acceleration or velocity signals. However, these methods suffer from inherent limitations, as their performance is strongly affected by factors such as sensor type~\cite{2022step_variation}, wear location~\cite{2017stepreview, sensorlocation2}, and variations in gait or stepping modes~\cite{imu1}. 
On the other hand, pressure sensors embedded in insoles~\cite{19reviewshoeinsole} or floors~\cite{17walkwaystepwearable} can capture ground reaction forces and the resulting pressure changes, providing reliable ground-truth measurements for step-level foot contact events. However, their applicability may be constrained in more natural settings. 

\vspace*{2mm}

\noindent \textbf{Step Counting in Computer Vision --}
Advances in human pose estimation have enabled its widespread use in gait analysis~\cite{22review_gaitanalysis}, which characterises walking patterns by measuring spatio-temporal gait features~\cite{21gaitanalysis,24gaitanalysis,2024visiongaitanalysis} such as stride length, swing time, and joint angles. For example, Connie et al.~\cite{gaitanalysis1} measure the relative distance between detected ankle keypoints as step length and estimate the number of steps by counting peaks in the resulting temporal signal. Similarly, Hii et al.~\cite{gaitanalysis2} detect heel-strike events as the time points at which the relative distance between the toe and hip joints reaches a maximum. However, these studies are primarily based on lab-based assessments with carefully scripted movements, whereas turning steps in real-world settings, especially for individuals with PD, are inherently diverse and aperiodic~\cite{2017stepreview, morgan2022understanding}, thereby violating the assumptions underlying existing gait analysis methods.
Alternatively, casting stepping as a primitive action reduces the task to repetitive activity counting (RAC)~\cite{repnet,ESCount,poserac}, which targets high-level activities with strong periodicity (e.g., jump rope), yet PD turning steps exhibit heterogeneous motion patterns with irregular temporal distribution. Our method shares the same spirit of an early method~\cite{05peak} that compress the  motion into 1D signal, whereas we similarly construct foot movement signals from human mesh recovery. Furthermore, we introduce a refine process where we use cross attention to query and retrieve detailed motion features from optical flow.

\begin{figure}[t]
    \centering
\includegraphics[width=\linewidth]{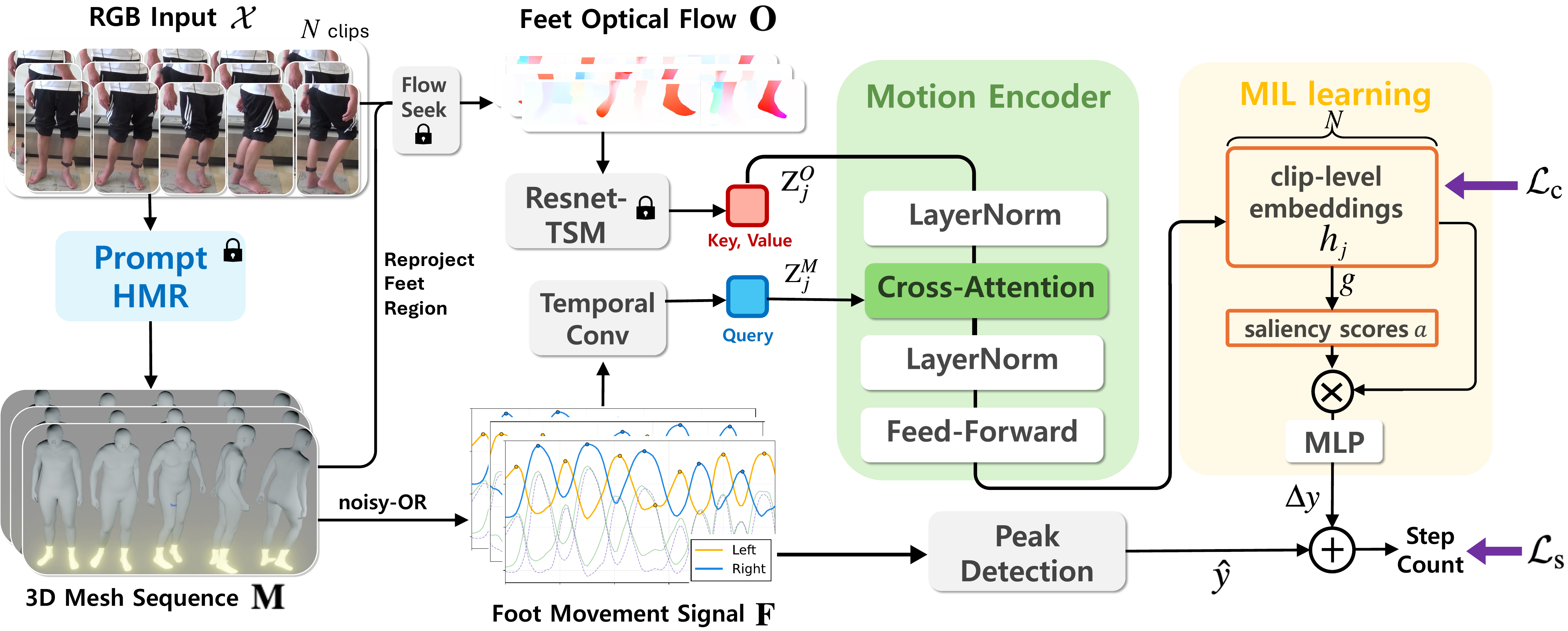}
          \caption{{Overall framework for step detection.} {A foot movement signal \textbf{F} from 3D human mesh recovery provides coarse step count estimation $\hat y$, where a cross attention motion encoder captures fine-grained motion cue by fusing mesh-based representation $\text{Z}_j^M$ and flow-based representation $\text{Z}_j^O$. We learn a clip-level saliency score to adaptively aggregate motion embeddings to predict a video-level residual $\Delta y$.}}
    \label{fig:main}
\end{figure}

\section{Methods} \label{sec: Methods}

In the proposed framework (see Figure~\ref{fig:main}),  we first obtain a coarse step count from a  foot movement signal derived from 3D human mesh recovery. 
To capture subtle foot dynamics beyond mesh recovery, we introduce a motion encoder that incorporates fine-grained motion cues from optical flow through cross attention.  
We divide videos into clips and combine the fused clip-level motion embeddings using MIL to predict a video-level residual for step count refinement.

\subsection{Foot Movement Signals from Human Mesh Recovery
}\label{sec:peakstep} 
First, we consider a step as an effective displacement of the foot that does not necessarily require full foot clearance, allowing the inclusion of clinically meaningful parkinsonian gait events such as shuffling or sliding steps. Then, given an input video $\mathcal{X}$ of turning
{with ${T}$ frames}, we reconstruct 
a 3D human mesh sequence $\mathbf{M}$ using {a human pose and shape} {estimation}  model (e.g., PromptHMR~\cite{prompthmr}), which also outputs per-joint stationary logits originally introduced for aligning global translation across frames. 
We repurpose the foot-related logits of $\mathbf{M}$ to construct a foot movement signal $\mathbf{F} = \{F_{i,t}\}$ (see Figure~\ref{fig:main}), indicating the {\it confidence} of whether {foot $i \in \{\text{left}, \text{right}\}$} is moving or stationary {at time $t \in [0,{T}-1]$.}
{Specifically, given the stationary logits
$\mathbf{r}=\{r_{i,t}^{\text{heel}}, r_{i,t}^{\text{toe}}\}$ for the toe and heel joints of both feet, movement probabilities $\mathbf{\pi}=\{\pi_{i,t}^{\text{heel}}, \pi_{i,t}^{\text{toe}}\}$  are computed as 
\begin{equation}
\pi_{i,t}^{\text{heel}}=\sigma(r_{i,t}^{\text{heel}}), \qquad
\pi_{i,t}^{\text{toe}}=\sigma(r_{i,t}^{\text{toe}}),
\end{equation}
where $\sigma(\cdot)$ denotes the sigmoid function.} 
Since a step should be counted when either part of the foot exhibits movement, we compute ${F}_{i,t}$ by fusing the toe and heel probabilities of each foot using a noisy-OR~\cite{nosiyor} operation,
\begin{equation}
\label{eq:noisyor}
{F}_{i,t} = 1 - (1 - \pi_{i,t}^{\text{heel}})(1 - \pi_{i,t}^{\text{toe}}).
\end{equation}
{Despite diverse PD turning patterns, each step follows a consistent cycle from static to moving to static; therefore, we obtain an initial step count estimate $\hat{y}$ as the number of peaks detected on $\mathbf{F}$, computed as 
\begin{equation}
\hat y = \sum_{i\in\{L,R\}} \sum_t p_{i,t},
\end{equation}} 
where $p_{i,t}$ denotes the soft peak response at time $t$ for foot $i$. We compute $p_{i,t}$ using a differentiable soft local-maximum head that compares each frame with its temporal neighbors
\begin{equation}
\label{eq:softpeak}
p_{i,t}
=
\sigma\!\left(
\beta({F}_{i,t}-{F}_{i,t-1})
\right)
\,
\sigma\!\left(
\beta({F}_{i,t}-{F}_{i,t+1})
\right),
\end{equation}
where $\beta$ controls peak sharpness. {A sensitivity analysis of $\beta$ is provided in the ablation study.}


\subsection{Flow-Assisted Step Count Refinement }\label{sec:refiner}

Although the mesh-based representation captures high-level structure of foot motion and provides an initial estimate of step count, recovering subtle and pathological foot movements remains challenging. To further refine the estimate, we fuse mesh-based and flow-based motion representations to learn clip-level motion embeddings, which are aggregated for video-level residual prediction.


\vspace*{2mm}

\noindent{\bf Clip-level Motion Encoder --  } 
As the temporal length $\mathcal{T}$ of input videos varies, we decompose $\mathcal{X}$ into $N$ non-overlapping temporal clips $\{x_{j}\}_{j=0}^{N-1}$ containing $l$ frames each, and use a shared motion encoder to fuse mesh-based and flow-based motion features to obtain clip-level embeddings $\{h_{j}\}_{j=0}^{N-1}$.
For the mesh branch, we treat the left and right foot movement signals as two channels and encode the raw motion signal using a temporal 1D convolution to model local temporal dependencies, yielding mesh-based motion tokens $\text{Z}_j^M \in \mathbb{R}^{l \times d}$, where $d$ is the token embedding dimension. 
For the flow branch, we extract dense optical flow \textbf{O} (using FlowSeek~\cite{poggi2025flowseek}) within the foot region of interest (RoI), obtained by projecting the reconstructed 3D mesh of feet~\cite{smpl} onto the image plane.
The raw optical flow is then encoded using ResNet-TSM~\cite{lin2019tsm} to capture spatial-temporal context, and subsequently projected to frame-level flow tokens $\text{Z}_j^O
\in \mathbb{R}^{l \times d}$. 
With the fine-grained motion preserved in per-pixel flow displacement field, we employ cross attention, where  $\text{Z}_j^M$ serves as queries over  $\text{Z}_j^O$ as keys and values to retrieve corresponding subtle foot movement details (see Figure~\ref{fig:main}).
Specifically,  we compute the embedding $h_j$ of the $j$-th clip as
\begin{equation}
\label{eq:attn_feat}
h_{j} =\text{Softmax}({QK}^\mathcal
{T}/\sqrt{d_k}){V},
\end{equation}
{
where 
\begin{equation}
Q = W_Q \,\mathrm{}\text{Z}_j^M, \quad
K = W_K \,\mathrm{}\text{Z}_j^O, \quad
V = W_V \,\mathrm{}\text{Z}_j^O, \quad
\end{equation}
and $W_Q,W_K,W_V\in\mathbb{R}^{d\times d}$ are learnable projection matrices.
}

\vspace*{2mm}

\noindent{\bf Video-level MIL Learning --} For each clip, cross attention enables $h_j$
to jointly reason over mesh-based and flow-based motion representations, either reinforcing consistent evidence shared by both representations or leveraging complementary cues from each other to compensate missing or spurious mesh-based steps. To adaptively aggregate the fused motion embeddings, we learn a saliency scoring function~\cite{ilse2018mil} $g$ for each clip and compute the video representation $\mathit{v}$ via weighted averaging
\begin{equation}
{v} = \sum_{j=0}^{N-1} a_j {h}_j, \quad a_j = \frac{\exp\left(g({h}_j)\right)}{\sum_{n=0}^{N-1} \exp\left(g({h}_n)\right)}.
\end{equation}
We regress step count offset $\Delta y$ using a prediction head over video-level embedding $v$ and the model is optimised with a squared error loss
\begin{equation}
\mathcal{L_\text{s}} = \left( {y} - (\hat y+\Delta y) \right)^2.
\end{equation}
During backpropagation, attention gradually shifts toward more informative clips, often corresponding to abnormal events (e.g., footskate) that cause counting inaccuracies, as higher-loss clips receive larger gradients that amplify their saliency scores 
\begin{equation}
\frac{\partial \mathcal{L_\text{s}}}{\partial a_j}
=
\frac{\partial \mathcal{L_\text{s}}}{\partial \Delta{y}}
\cdot
\frac{\partial \Delta{y}}{\partial \mathit{v}}
\cdot
\frac{\partial \mathit{v}}{\partial a_j}
\propto
h_j.
\end{equation}
To better separate reliable and error-prone clips, we also introduce a contrastive regularisation term similar to InfoNCE~\cite{MOCO}, applied to the top-$k$ and bottom-$k$ clip embeddings~\cite{zhang2021cola} ranked by saliency scores $a_j$, such that 
\begin{equation}
\mathcal{L}_{\text{c}} =
\sum_{u}
-\log
\frac{
\sum_{j \in u} \exp(h_u^\top h_j)
}{
\sum_{j} \exp(h_u^\top h_j) 
},
\end{equation}
where $u \in \{\text{top-}k,\text{bottom-}k\}$ and $h_u^\top$ is the anchor embedding obtained using mean pooling.
Finally, the total loss is 
\begin{equation}
\label{eq:loss}
\mathcal{L} = \mathcal{L_\text{s}}+ \lambda \mathcal{L_\text{c}},
\end{equation}
where $\lambda$ is the hyperparameter balancing these two objectives. We ablate the
value $\lambda$ in Section~\ref{sec:ablations} to study its influence on performance. 
\section{Experiments} \label{sec: Experiments}
\subsection{Experimental Setup}
\textbf{Datasets --}
We validate our step counting method on two real-world PD video datasets (see Table~\ref{tab:dataset}), namely PD-FOG~\cite{PDFOG} and {Turn-REMAP}~\cite{remap}. 
PD-FOG contains 52 {long-duration} videos of 25 {PD patients} instructed to perform turning-in-place tasks at their own normal speed in clinical settings, while wearing a 6-axis IMU on one ankle. {We exclude videos without turning behavior due to severe freezing (e.g., medication-induced foot trembling~\cite{fog}). Though clinically valuable, freezing is a distinct action category that is not meaningfully measured by step counting.}
Turn-REMAP contains 3729 trimmed turning videos collected during daily activities from 11 pairs of participants with PD and healthy controls (HCs), during unobstructed free living in a real home environment.  Neither dataset includes people with very severe PD for whom turning may be unsafe due to the high risk of falling. The ground-truth step counts for both datasets were annotated by two trained experts, with 50\% of the annotations cross-checked by both annotators. 

\begin{table}[t]
\centering
\caption{{PD turning datasets characteristics.} {PD-FoG contains clinically controlled turning-in-place tasks, while Turn-REMAP captures natural turning behaviour during free-living activities in real-world settings.}}
\label{tab:dataset}

\resizebox{0.8\textwidth}{!}{
\begin{tabular}{lrrrrr}
\toprule
\textbf{Dataset} & \textbf{\# videos} & \textbf{\# HC} & \textbf{\# PD} & \textbf{Avg. duration} & \textbf{Avg. \# steps} \\
\midrule

PD-FoG & 52 & 0 & 25 & 118.2s & 284.3 \\

Turn-REMAP & 2477 & 11 &11 & 1.88s &2.47\\

\bottomrule
\end{tabular}
}
\end{table}

\vspace*{2mm}

\noindent \textbf{Evaluation Metrics --} 
For evaluation, we report the mean absolute error (MAE) between the predicted and ground truth step counts as a direct measure of prediction error, as well as accuracy {under relative error thresholds of 5\%, 10\% and 20\% of the total number of steps, denoted as $\text{ACC}_{95}$, $\text{ACC}_{90}$ and $\text{ACC}_{80}$.}
We use leave-one-subject-out (LOSO) cross-validation~\cite{loso} on the PD-FOG dataset due to the limited number of subjects in the turning subset. For Turn-REMAP, healthy subjects 1022 and 1027, together with PD subjects 1030 and 1032, are used for validation, while the remaining subjects are used for training.

\vspace*{2mm}

\noindent \textbf{Compared Methods --}  We compare our proposed method against three categories of step-counting approaches: (i) IMU-based methods~\cite{imu1,imu2,imu3}, which estimate step counts from acceleration and angular velocity signals measured by foot-mounted sensors.
(ii) Gait-analysis methods~\cite{gaitanalysis1,gaitanalysis2}, which use pose estimation to derive biomechanical measurements from which step counts can be inferred. (iii) Repetitive activity counting (RAC) methods~\cite{repnet,ESCount}, which design end-to-end networks to learn temporal patterns directly from video data. 



{All methods are evaluated using the same train/val split for both dataset. Learnable models, such as RepNet~\cite{repnet}, are trained from scratch, while non-trainable baselines, such as IMU-based methods~\cite{imu1,imu3}, use the training split for algorithmic parameter selection (e.g., detection thresholds and smoothing windows).}

\vspace*{2mm}

\noindent \textbf{Implementation Details --} 
We use PromptHMR~\cite{prompthmr} for 3D human mesh recovery, FlowSeek ~\cite{poggi2025flowseek} for optical flow extraction, and a K400-pretrained ResNet-TSM~\cite{lin2019tsm} backbone for flow feature encoding; all upstream modules are kept frozen during training.
Input videos are uniformly partitioned into non-overlapping clips of 30 frames. Our motion encoder consists of two cross attention layers with the token dimension $d$ = 128. The model is optimised using SGD with momentum 0.9, weight decay 0.001 and a learning rate of 0.0001. The contrastive loss weight $\lambda$ is set to 0.05 and the top- and bottom-$k$ clips are selected from the top and bottom 25\% of clips, respectively.

\subsection{Parkinsonian Step Counting in Real World}
\begin{table}[t]
\centering
\caption{
Step counting performance on PD-FoG. Mild and moderate groups are defined based on the standard H\&Y~\cite{h&y} disease severity score.
} 
\label{tab:results_step_fog}
\setlength{\tabcolsep}{2pt}
\renewcommand{\arraystretch}{1.35}
\resizebox{\textwidth}{!}{
\begin{tabular}{clrcccrccc|rccc}
\toprule

 & & \multicolumn{4}{c}{\textbf{Mild}}
 & \multicolumn{4}{c}{\textbf{Moderate}}
 & \multicolumn{4}{|c}{\textbf{All}} \\

\cmidrule(lr){3-6} 
\cmidrule(lr){7-10} 
\cmidrule(lr){11-14}

 & \textbf{Methods} 
 & MAE$\downarrow$ & $\text{ACC}_{95}\uparrow$ & $\text{ACC}_{90}\uparrow$ & $\text{ACC}_{80}\uparrow$
 & MAE$\downarrow$ & $\text{ACC}_{95}\uparrow$ & $\text{ACC}_{90}\uparrow$ & $\text{ACC}_{80}\uparrow$
 & MAE$\downarrow$ & $\text{ACC}_{95}\uparrow$ & $\text{ACC}_{90}\uparrow$ & $\text{ACC}_{80}\uparrow$ \\

\cmidrule{1-14}

\multirow{3}{1em}{\rotatebox[origin=c]{90}{\textbf{IMU}}}

& Liu et al.~\cite{imu1}   
& 179.6 & 0.024 & 0.024 & 0.146
& 180.2 & 0.000 & 0.000 & 0.000 
& 179.7 & 0.019 & 0.019 & 0.115 \\

& Bi et al.~\cite{imu3}    
& 57.2  & 0.244 & 0.415 & 0.615
& 54.5  & \underline{0.273} & \underline{0.455} & \underline{0.585}
& 56.6  & 0.250 & 0.423 & 0.615 \\

& Lucot et al.~\cite{imu2} 
& \underline{30.6} & \underline{0.488} & \underline{0.585} & \underline{0.854}
& {64.4} & 0.091 & 0.364 & {0.546}
& \underline{37.7} & \underline{0.404} & \underline{0.538} & \underline{0.779} \\

\midrule

\multirow{2}{1em}{\rotatebox[origin=c]{90}{\textbf{pose}}}

& Connie et al.~\cite{gaitanalysis1} 
& 53.4 & 0.220 & 0.390 & 0.610
& 56.2 & 0.182 & 0.273 & 0.455
& 54.0 & 0.212 & 0.365 & 0.576 \\

& Hii et al.~\cite{gaitanalysis2}    
& 65.5 & 0.146 & 0.268 & 0.577
& 80.6 & 0.000 & 0.182 & 0.364
& 68.7 & 0.115 & 0.250 & 0.634 \\

\midrule

\multirow{2}{1em}{\rotatebox[origin=c]{90}{\textbf{RAC}}}

& RepNet~\cite{repnet}   
& 62.2 & 0.220 & 0.317 & 0.634
& 61.8 & \underline{0.273} & \underline{0.455} & {0.546}
& 62.2 & 0.231 & 0.346 & 0.615 \\

& ESCount~\cite{ESCount} 
& 47.2 & 0.244 & 0.366 & 0.610
& \underline{45.5} & 0.091 & 0.273 & 0.524
& 46.8 & 0.212 & 0.346 & 0.634 \\

\midrule

& Ours
& \textbf{13.1} & \textbf{0.634} & \textbf{0.927} & \textbf{0.951}
& \textbf{18.4} & \textbf{0.546} & \textbf{0.818} & \textbf{1.000}
& \textbf{15.9} & \textbf{0.615} & \textbf{0.904} & \textbf{0.962} \\

\bottomrule
\end{tabular}
}
\end{table}

\noindent \textbf{Results on PD-FOG -- } \label{sec: results_PDFOG} We group PD-FOG subjects by Hoehn \& Yahr (H\&Y) severity score~\cite{h&y} into mild (1–2) and moderate (2.5–3) groups. 
As shown in Table~\ref{tab:results_step_fog}, our method consistently outperforms existing 
{IMU-based and adapted vision-based methods for step counting}
on PD-FOG by a large margin across different severity levels. For example, compared to the IMU-based method of Lucot et al.~\cite{imu2}, we reduce the absolute error from 37.7 to 15.9 and improve the accuracy at a 20\% error  threshold from 0.779 to 0.962 (+18.3\%). Similar gains are observed over the pose-based gait analysis approach of Connie et al.~\cite{gaitanalysis1}, where the error decreases from 54.0 to 15.9 and the corresponding accuracy improves from 0.576 to 0.962 (+38.6\%). 
{Our method consistently outperforms RAC-based methods such as RepNet~\cite{repnet} and ESCount~\cite{ESCount}, reducing the MAE from 62.2 and 46.8 to 15.9, respectively. Correspondingly, the accuracy improves to 0.962 from 0.615 (+34.7\%) and 0.634 (+32.8\%).}
Notably, our method yields a lower step count error on participants with moderate PD severity, as their stepping patterns are typically more irregular and subtle than those of mild cases.

\begin{table}[t]
\centering
\caption{Step counting performance on Turn-REMAP. Due to the relatively small step counts in this dataset, MAE are reported to three decimal places.}
\label{tab:results_step_remap}
\setlength{\tabcolsep}{2pt}
\renewcommand{\arraystretch}{1.35}
\resizebox{\textwidth}{!}{
\begin{tabular}{clrcccrccc|rccc}
\toprule

 & & \multicolumn{4}{c}{\textbf{PD}}
 & \multicolumn{4}{c}{\textbf{HC}}
 & \multicolumn{4}{|c}{\textbf{All}} \\

\cmidrule(lr){3-6} 
\cmidrule(lr){7-10} 
\cmidrule(lr){11-14}

 & \textbf{Methods} 
 & MAE$\downarrow$ & $\text{ACC}_{95}\uparrow$ & $\text{ACC}_{90}\uparrow$ & $\text{ACC}_{80}\uparrow$
 & MAE$\downarrow$ & $\text{ACC}_{95}\uparrow$ & $\text{ACC}_{90}\uparrow$ & $\text{ACC}_{80}\uparrow$
 & MAE$\downarrow$ & $\text{ACC}_{95}\uparrow$ & $\text{ACC}_{90}\uparrow$ & $\text{ACC}_{80}\uparrow$ \\

\cmidrule{1-14}
\multirow{2}{1em}{\rotatebox[origin=c]{90}{\textbf{pose}}}

& Connie et al.~\cite{gaitanalysis1} 
& 1.058 & 0.022 & 0.216 & 0.295
& 0.899 & 0.000 & 0.221 & 0.332
& 0.986 & 0.012 & 0.219 & 0.312 \\

& Hii et al.~\cite{gaitanalysis2}   
& 0.904 & 0.019 & 0.198 & 0.347
& \underline{0.651} & 0.013 & \textbf{0.252} & \underline{0.438}
& 0.789 & 0.016 & \underline{0.223} & 0.389 \\

\midrule

\multirow{2}{1em}{\rotatebox[origin=c]{90}{\textbf{RAC}}}
& RepNet~\cite{repnet}
& 1.431 & 0.022 & 0.045 & 0.097
& 1.112 & 0.035 & 0.071 & 0.133
& 1.286 & 0.028 & 0.057 & 0.113

\\

& ESCount~\cite{ESCount} 
&\underline{0.634} & \underline{0.071}&\underline{0.228}&\textbf{0.474}
&{0.653} & \underline{0.084}&0.208&\textbf{0.447}
&\underline{0.643}&  \underline{0.077}&0.219 &\textbf{0.462}
\\
\midrule

& Ours
& \textbf{0.615} & \textbf{0.231} & \textbf{0.343} & \underline{0.470}
& \textbf{0.637} & \textbf{0.142} & \underline{0.239} & {0.412}
& \textbf{0.625} & \textbf{0.190} & \textbf{0.296} & \underline{0.443} \\

\bottomrule
\end{tabular}
}
\end{table}

\vspace*{2mm}
\noindent \textbf{Results on Turn-REMAP -- } \label{sec: results_TurnREMAP}
Our method again {generally} outperforms repurposed vision-based methods 
(see Table~\ref{tab:results_step_remap}). For example, compared to Hii et al.~\cite{gaitanalysis2}, the {mean} absolute step counting error is reduced from 0.789 to 0.625, while accuracy is improved from 0.389 to 0.443 (+5.4\%).  In particular, our method achieves a substantial gain in accuracy at a 5\% error threshold, from 0.077 to 0.190 (+11.3\%) over the second-best method, ESCount~\cite{ESCount}. The only exception is accuracy at the 20\% error threshold, where our method performs slightly worse than ESCount, decreasing from 0.462 to 0.443 (-1.9\%).

\begin{table}[h!]
\centering
\caption{{Ablations on the proposed designs.}}
\label{tab:ablations_design}
\setlength{\tabcolsep}{5pt}
\resizebox{0.7\textwidth}{!}{
\begin{tabular}{l rrrrr}
\toprule

\multirow{1}{*}{\textbf{Settings}}  
& \textbf{MAE} $\downarrow$ 
& \textbf{$\text{ACC}_{95}\uparrow$} 
& \textbf{$\text{ACC}_{90} \uparrow$}
& \textbf{$\text{ACC}_{80} \uparrow$} \\

\midrule

$\textit{w/o}$ cross attention 
& 17.5 
& \underline{0.596} 
& 0.846 
& \textbf{0.962} \\

$\textit{w/o}$ attentional pooling 
& 17.3 
& 0.577 
& 0.846  
& \textbf{0.962} \\

$\textit{w/o}$ $\mathcal{L}_c$ ($\lambda = 0$) 
& \underline{16.0} 
& \textbf{0.615}
& \underline{0.865} 
& \textbf{0.962} \\

\midrule
ours 
& \textbf{15.9} 
& \textbf{0.615} 
& \textbf{0.904}
& \textbf{0.962} \\

\bottomrule
\end{tabular}
}
\end{table}
\subsection{Ablations}
\label{sec:ablations}
\textbf{Ablations on Proposed Designs -- }To understand the impact of each component in
our method, we evaluate three variants of our model using the same training and evaluation protocol on the PD-FOG dataset. The results are shown in Table~\ref{tab:ablations_design}: (1) $\textit{w/o}$ cross attention: instead of using mesh features to query flow features, we concatenate them as input tokens and apply self-attention. The performance drop suggests that cross attention introduces a beneficial inductive bias by enforcing directed interactions between coarse mesh features and fine-grained flow features. 
(2) $\textit{w/o}$ attentional pooling: substituting the learnable attentional pooling with to max pooling leads to worse results, highlighting the effectiveness of attention in capturing informative local cues. (3) $\textit{w/o}$ $\mathcal{L}_c$: omitting the contrastive regularisation slightly reduces performance, suggesting that it encourages a more discriminative clip-level embedding space.

\vspace*{2mm}

\noindent \textbf{Ablations on $\lambda$ -- }{We study the effect of $\lambda$, the weighting coefficient of the contrastive loss $\mathcal{L}_c$ in Eq.~(\ref{eq:loss}).} As $\lambda$ increases (see Table~\ref{tab:ablations_lambda}), the model becomes overly regularised, {prioritising} the separation of reliable and error-prone clips over informative representation learning. We use $\lambda=0.05$ as the default setting in all experiments. 
\begin{table}[t]
\centering
\caption{{Ablations on $\lambda$. }}
\label{tab:ablations_lambda}
\setlength{\tabcolsep}{5pt}
\resizebox{0.6\textwidth}{!}{
\begin{tabular}{l rrrrr}
\toprule

\multirow{1}{*}{\textbf{Settings}}  
& \textbf{MAE} $\downarrow$ 
& \textbf{$\text{ACC}_{95}\uparrow$} 
& \textbf{$\text{ACC}_{90} \uparrow$}
& \textbf{$\text{ACC}_{80} \uparrow$} \\

\midrule

$\lambda = 0.01$ & \underline{16.0} &\underline{0.577}&\underline{0.885}&\textbf{0.962}\\
$\lambda = 0.05$ & \textbf{15.9} 
& \textbf{0.615} 
& \textbf{0.904}
& \textbf{0.962} \\
$\lambda = 0.1$ & 16.9 &0.558&\underline{0.885}&\textbf{0.962}\\
$\lambda = 0.2$ & 16.6 &\underline{0.577}&{0.865}&\textbf{0.962}\\

\bottomrule
\end{tabular}
}
\end{table}

\vspace*{2mm}

\noindent \textbf{Ablations on Multi-modal Motion Feature  -- } In Table~\ref{tab:ablate_modality}, we show ablations on additional multi-modal motion representations by incorporating RGB and/or IMU features into the default mesh \& flow setting.
Specifically, RGB features are extracted using the same ResNet-TSM backbone, while 6-channel IMU signals are encoded with a 1D convolutional network. These features are concatenated with flow features and projected into the token dimension as keys and queries.
Adding RGB features does not improve MAE, suggesting that appearance information provides limited complementary cues beyond the motion-aware flow features. Incorporating IMU signals also fails to improve performance, possibly because the relevant motion cues are subtle and easily entangled with incidental body movements. Combining RGB and IMU features also provides no further benefit.



\begin{table}[h]
\centering
\caption{Ablations on multi-modal feature input. We investigate the integration of additional RGB and IMU modalities into our default mesh \& flow setting. }
\label{tab:ablate_modality}
\setlength{\tabcolsep}{5pt}
\resizebox{0.7\textwidth}{!}{
\begin{tabular}{l rrrr}
\toprule

\multirow{1}{*}{\textbf{Settings}}  
& \textbf{MAE$\downarrow$}  
& \textbf{$\text{ACC}_{95}\uparrow$} 
& \textbf{$\text{ACC}_{90} \uparrow$}
& \textbf{$\text{ACC}_{80} \uparrow$} \\

\midrule

mesh \& flow (default) 
& \textbf{15.9} 
& \textbf{0.615} 
& \textbf{0.904}
&  \textbf{0.962}\\

\midrule

+ RGB 
& \underline{17.4}
& \underline{0.596}
& 0.865
& \textbf{0.962} \\

+ IMU  
& 18.3 
& 0.560
& 0.860
& \underline{0.960} \\

+ RGB \& IMU 
& 18.6
& {0.460}
& \underline{0.900}
& \underline{0.960} \\

\bottomrule
\end{tabular}
}
\end{table}

\begin{table}[h]
\centering
\caption{Sensitivity analysis on $\beta$ on PD-FOG.}
\label{tab:ablate_beta}
\setlength{\tabcolsep}{5pt}

\begin{tabular}{l ccccc}
\toprule

$\beta$ (sharpness of peaks) &1e4&5e4&1e5&5e5&1e6\\
\midrule
$\hat y$ (detected peaks) &297.23&293.64&293.50&293.41&293.41\\

\bottomrule
\end{tabular}

\end{table}

\vspace*{2mm}

\noindent \textbf{Ablations on $\beta$ -- }We study the effect of $\beta$ in Eq.~(\ref{eq:softpeak}), which controls detector sharpness, on the initial coarse step-count estimate $\hat y$. As shown in Table~\ref{tab:ablate_beta}, the predicted peak count becomes insensitive to $\beta$ for sufficiently large values, where the detector closely approximates hard peak counting while remaining differentiable.

\begin{figure}[t!]
    \centering
     \resizebox{0.82\textwidth}{!}{\includegraphics[width=0.5\linewidth]{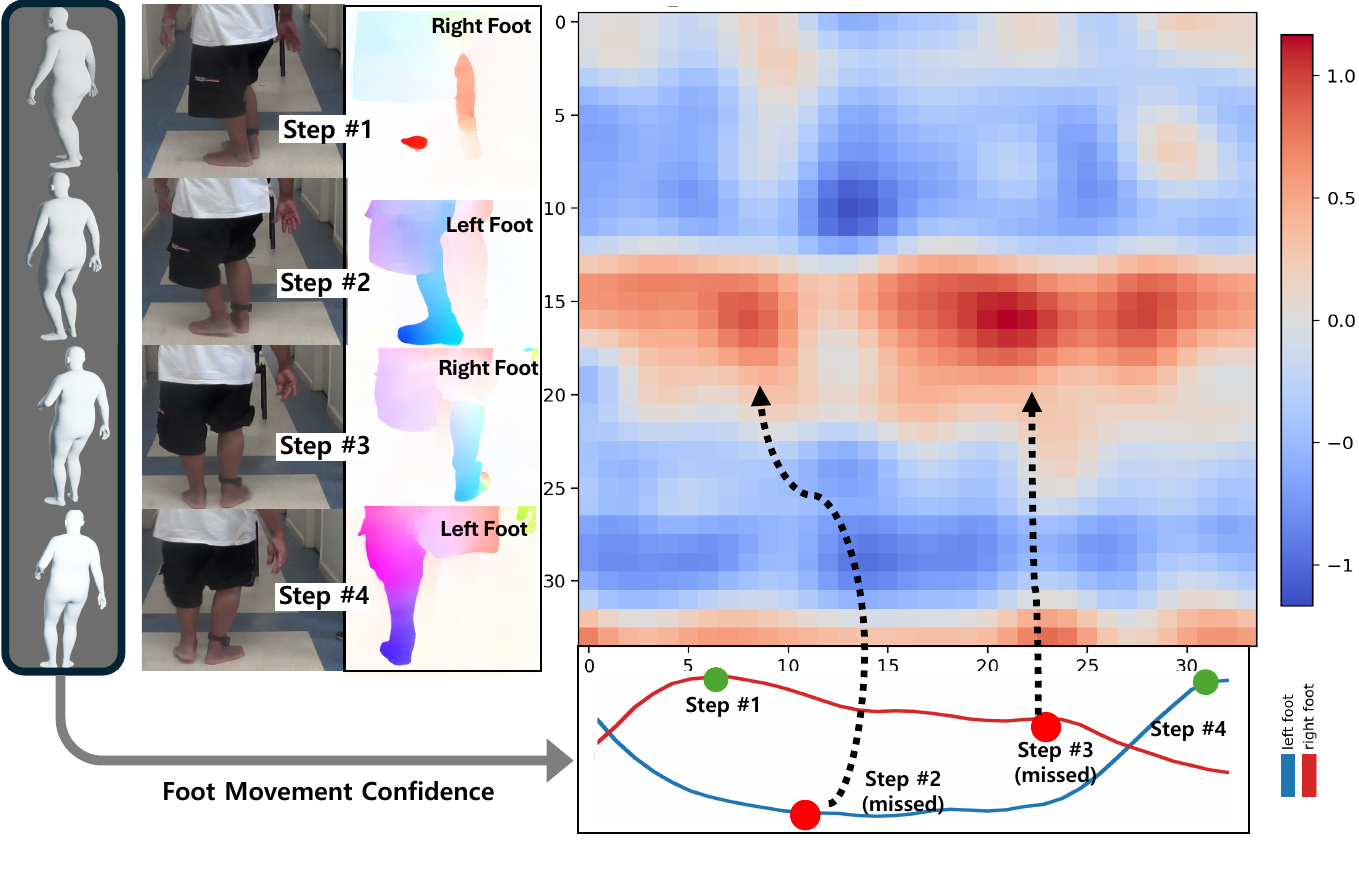}  }
    \caption{{Cross attention in motion encoder.}  The horizontal axis corresponds to mesh-based foot movement queries and the vertical axis corresponds to optical flow key-value tokens. The encoder is able to reason over inconsistencies between mesh-based foot motion and optical flow, where the second and third step cycles are missed by the mesh signal but preserved in the optical flow. 
    }
    \label{fig:cross_attn}
\end{figure}

\begin{figure}[b!]
    \centering
     \resizebox{0.82\textwidth}{!}{\includegraphics[width=0.5\linewidth]{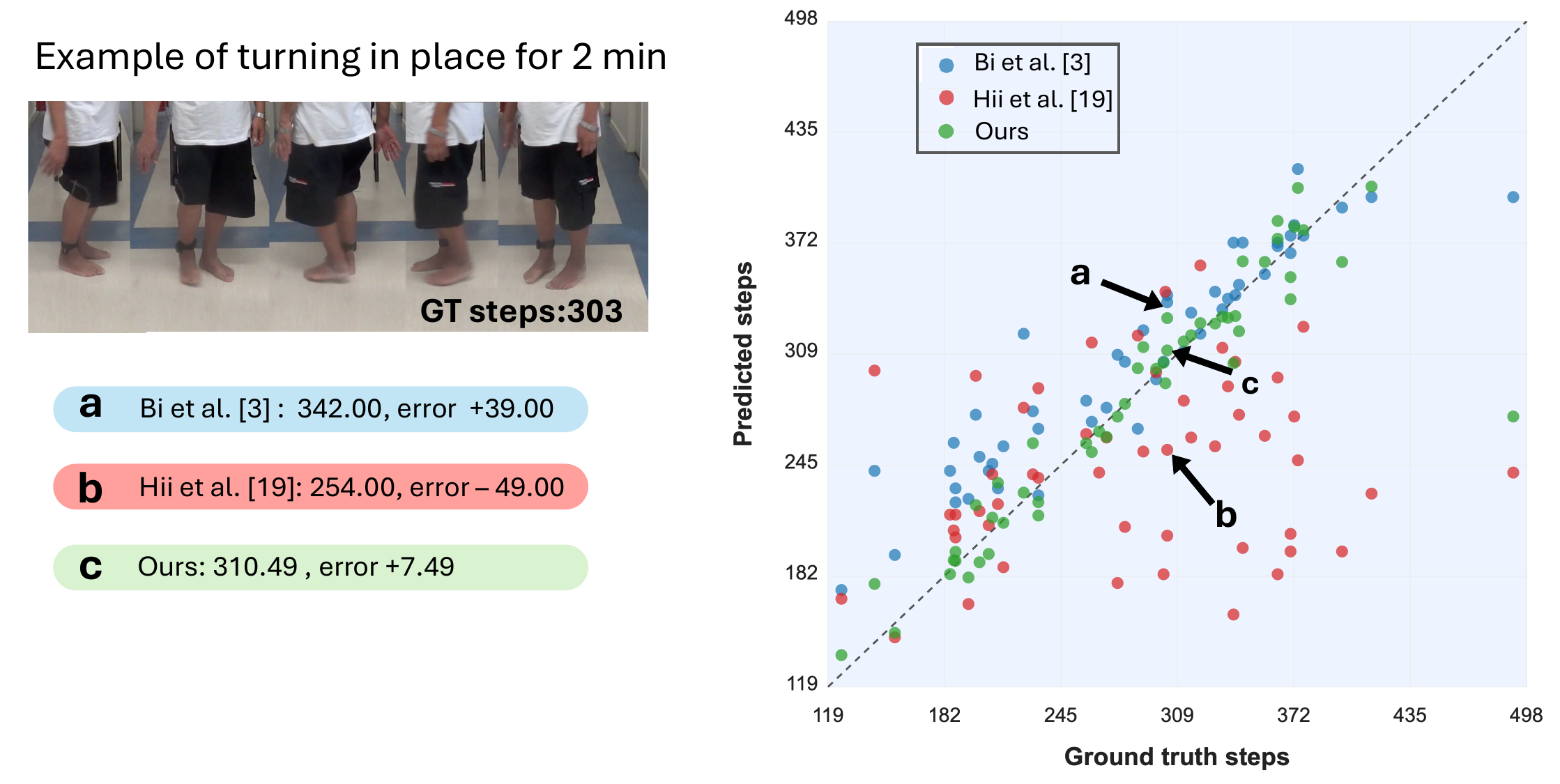}  }
    \caption{Step count estimation results on PD-FOG. (Left) Frames from an example from clip {PDFE30\_2} with 303 steps, where our method achieves lower counting error than prior methods;  (Right) Predicted versus ground-truth step counts for Bi et al.~\cite{imu3}, Hii et al.~\cite{gaitanalysis2}, and our method.}
    \label{fig: comparison}
\end{figure}

\subsection{Qualitative Results}
In Figure~\ref{fig:cross_attn}, {we visualise the cross attention map of the last layer of motion encoder for an example containing four small turning steps, following a \textit{right-foot/left-foot/right-foot/left-foot} stepping sequence.} The mesh representation only captures the $1^{st}$ and last steps, while the $2^{nd}$ and $3^{rd}$ steps are smoothed out as footskate, whereas these subtle movements are preserved in the optical flow. The attention focuses on cross-modal inconsistencies, encoding complementary cues for subsequent residual step count refinement in video-level MIL.

{In Figure~\ref{fig: comparison} (right), we illustrate that our method's predictions cluster more closely around the ground-truth (dashed diagonal line) and thus yield more accurate step count estimation across subjects compared to prior methods~\cite{imu3,gaitanalysis2}. In Figure~\ref{fig: comparison} (left), frames from a specific example case are shown, alongside the groundtruth count. 
Prior methods exhibit substantial overcounting or undercounting, whereas our method achieves the smallest count error}. 

\begin{figure}[t]
    \centering
     \resizebox{0.98\textwidth}{!}{\includegraphics[width=0.5\linewidth]{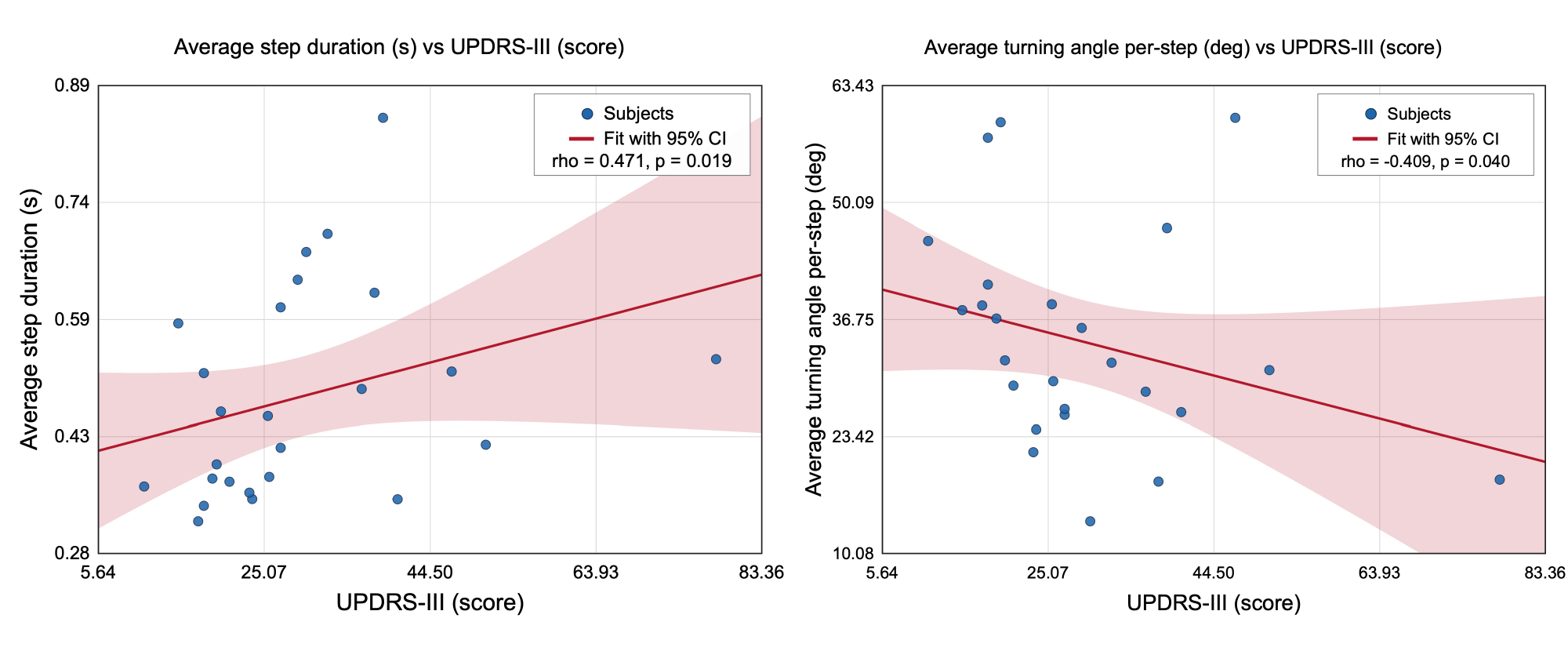}  }
    \caption{Illustration of the correlations between the step-level turning parameters and the MDS-UPDRS III score.}
    \label{fig: corr}
\end{figure}

\subsection{Clinical Correlation and Statistical Significance}
To demonstrate how our step count estimation can be used to calculate step-level turning parameters to quantify PD motor impairment, we perform a correlation analysis against the gold-standard MDS-UPDRS Part III motor score using Spearman's $\rho$ and $p$-values on PD-FOG dataset, which is shown in Figure~\ref{fig: corr}. Average step duration, computed as the duration divided by the number of steps, shows a statistically significant positive correlation with motor symptom severity ($\rho=0.471$, $p=0.019$). In contrast, average turning angle per-step, computed as the total turning angle divided by the number of steps, exhibits a statistically significant negative correlation ($\rho=-0.409$, $p=0.040$).
These findings suggest that, as the disease progresses, people with PD require more time to initiate steps and exhibit smaller angular displacements per step during turning-in-place assessments. {By enabling quantitative, step-level assessment of turning impairment from real-world videos, our approach can provide accessible and sensitive digital biomarkers to support earlier detection of PD, more personalised symptom tracking, and objective endpoints in clinical trials.} 

\section{Conclusion}
In this work, we introduce a passive video-based framework for automatically counting turning steps in people with PD. To capture the shuffled, subtle, and heterogeneous turning stepping patterns in PD, we investigate two commonly used motion representations, namely 3D human mesh and optical flow, and design a cross attention feature fusion encoder that uses coarse mesh-based signals as query tokens to retrieve fine-grained motion cues from optical flow. Informative clip-wise motion embeddings are selectively aggregated through a MIL-based strategy for step count residual prediction. Extensive experiments on both in-clinic and at-home free-living PD turning datasets demonstrate superior performance over existing vision-based and IMU-based approaches. Our method provides a practical alternative to wearable-based solutions with broader applicability, enabling more accurate quantification of PD turning abnormalities through step counting.


%
\bibliography{egbib}
\end{document}